# Learning to Navigate the Energy Landscape

Julien Valentin[1,3]   Angela Dai[2]   Matthias Nießner[2]
Pushmeet Kohli[1]   Philip Torr[3]   Shahram Izadi[1]   Cem Keskin[1]
[1]Microsoft Research   [2]Stanford University   [3]University of Oxford

## Abstract

*In this paper[1], we present a novel, general, and efficient architecture for addressing computer vision problems that are approached from an 'Analysis by Synthesis' standpoint. Analysis by synthesis involves the minimization of reconstruction error, which is typically a non-convex function of the latent target variables. State-of-the-art methods adopt a hybrid scheme where discriminatively trained predictors like Random Forests or Convolutional Neural Networks are used to initialize local search algorithms. While these hybrid methods have been shown to produce promising results, they often get stuck in local optima. Our method goes beyond the conventional hybrid architecture by not only proposing multiple accurate initial solutions but by also defining a navigational structure over the solution space that can be used for extremely efficient gradient-free local search. We demonstrate the efficacy and generalizability of our approach by on tasks as diverse as Hand Pose Estimation, RGB Camera Relocalization, and Image Retrieval.*

## 1. Introduction

Recent years have seen the re-emergence of the 'analysis by synthesis' or 'inverse graphics' approach to Computer Vision problems [3, 41, 38, 26, 20, 42, 43]. This elegant technique works by finding the parameters of the synthesis model that minimizes the reconstruction error; e.g., the distance between synthesized and query images. It has been explored by researchers many times [16, 4, 45, 49] but repeatedly fell out of favor due to the perception of being computationally expensive and brittle due to reconstruction error being a non-convex function of the target variables [25].

More lately, however, the availability of large datasets of training data along with the ability to train high capacity models like Convolutional Networks (CNN) [44, 26] and Random Forests (RF) [41, 38] have given the approach a new breath of life. These models have led to the development of hybrid architectures, where discriminatively trained feed-forward predictors (RFs or CNNs) are used to initialize continuous local search algorithms. While these hybrid techniques have been shown to produce impressive results, it is hard to make them work robustly in real time. This is in large part due to the cost of computing the derivative of the reconstruction error with respect to the model parameters, as it is a function of the rendering process. Although gradient-free optimization methods such as Particle Swarm Optimization (PSO) exist, they suffer from the problem that generative methods often require a significant amount of error evaluations to converge to good solutions. This can be very expensive, as each error evaluation requires performing a rendering operation on-the-fly[2]. Another major obstacle is that it is non-trivial to extract good search seeds in super real-time. This is a requirement for any continuous or stochastic optimization operating on a non-convex surface to converge to good minima while allowing the entire optimization pipeline to run in real-time. There is then a need for flexible and general optimization frameworks mimicking the generative optimization procedure, but orders of magnitude faster than the current methods.

In this paper, we propose a new general framework for minimizing the reconstruction error emanating from the analysis by synthesis approach. Our approach is based on integration of two search structures: retrieval forests and a navigation graph. The retrieval forest is composed of a collection of trees, each of which operates like a learned, conditional, locality sensitive hash function that maps to a hash table bucket (leaf) where a subset of the instances from the training set are stored which can be used as candidates for local search. However, instead of computing the derivative of the rendering pipeline or doing a blind search like PSO, our local search method operates by traversing a hierarchical navigation graph defined on the parameter space. Each instance in the training set has a corresponding vertex in the graph which is connected with a set of neighboring vertices (ideally i.i.d. distributed in parameter space around that vertex) according to a distance measure. A simple choice for the distance for defining the neighborhood is the L1- or L2-distance in descriptor-space, as it is involved in the computation of the reconstruction error, but L2-distance in target parameter space can also be used to build the graph.

---

[1] All of our datasets are publicly available (for details, see Sec. 7.3.): http://www.graphics.stanford.edu/projects/reloc/

[2] It is worth noting that it is sometimes possible to alleviate those limitations by heavy code engineering [38] to render small images.

Traversal of the edges in this navigational graph can be seen as making updates in the parameter space in the rough direction of the gradient minimizing the reconstruction error. We build this navigational graph with multiple levels, where vertices in higher levels are more sparse, allowing for larger jumps across the parameter space.

During test time, we use a collection of discriminatively trained retrieval trees to propose candidate vertices from where the local search should be initiated. We then repeatedly consider all the neighbors of the candidate vertices, compute the distance between them and the query image (reconstruction error or surrogate error), and select those having lowest error as candidates for future exploration. We consider a Navigation Graph to be converged when no new vertices with a better energy can be reached. Then, we initiate a new search using the next Graph in the hierarchy, and proceed until the convergence of all the Navigation Graphs in the hierarchy. At this stage, the best prediction can be directly used as a prediction. Alternatively, the most promising predictions can be used to seed a continuous (e.g., Levenberg-Marquardt) or stochastic procedure (e.g., Particle Swarm Optimization) for a few steps of local and final refinement.

We first demonstrate the efficacy and generalizability of our approach on Hand Pose Estimation and Image Retrieval tasks. We also demonstrate the efficacy of our approach on the problem of RGB Camera Relocalization. To make this problem particularly challenging, we introduce a new dataset of 3D environments that are significantly larger than those found in other publicly-available datasets. Our experimental results indicate that the composite retrieval tree-navigation graph architecture not only leads to dramatic improvements in computation time but also results in more accurate solutions.

To summarize, the main contributions of this work are (i) a new gradient-free heuristic optimization method based on navigational graphs that is fast and accurate, (ii) the introduction of a new public dataset for RGB and RGB-D relocalization that is significantly larger that those currently available, (iii) the demonstration that our approach is generic and reaches state-of-the-art results on three very distinct vision problems, namely RGB camera relocalization, hand pose estimation, and image retrieval.

## 2. Related work

We briefly introduce popular methods for approximate nearest neighbor search in the next four paragraphs. We refer the interested reader to [15, 14, 24, 10, 48, 36, 47] and [41, 38] for relevant litterature on RGB camera relocalization and Hand Pose Estimation, respectively.

**LSH** A number of papers in the literature have considered the use of hash functions for solving regression problems such as human pose estimation [37]. These are based on the concept of locality sensitive hashing (LSH) [19], using several random projection functions to hash each point of the dataset so that similar items map to the same bucket with high probability. Our approach is related to the pose-sensitive hashing method [37], as they focus on retrieving neighbors in the latent low-dimensional manifold where the data lives (articulated human pose) rather than the observed space of images. In this paper, the main target is to minimize the camera pose distance between the predictions and the actual poses of the 'lost' cameras.

In their influential work, the authors of [34] show how to use simple approximations that allow better probing sequences for LSH. Their multiprobe LSH led to improvements in both space and time. Significant work on building efficient index that can be used by LSH include [50] where the authors propose a compounds representation by decomposing images into background and image-specific information. For additional details on LSH, see [33, 1, 34].

**KD-tree** Contrary to LSH, a KD-tree [5, 12] recursively partitions the space with axis-aligned hyperplanes that split the data in two halves. Notable work in this area include [32] where the authors build a hierarchical $k$-means clustering and the widely used library based on [28] More details about different variants of KD-tree can be found in [7].

**Derivative-free optimization** Particle Swarm Optimization [23], Cuckoo Search, and Genetic Algorithms are search heuristics. The common idea behind these methods is to perturbate the current solution set (random i.i.d. jumps, mutations, etc.) in the hope of finding a new solution set that better explains the target. In the context of hand pose estimation, such methods can be used [18] to iteratively refine the pose parameters that best explain the input image. We refer the reader to [8] for extensive discussions on derivative-free optimization.

$k$**-NN graphs** There are a number of works that focus on hill-climbing strategies or $k$-NN graphs, but to the best of our knowledge there has been only one attempt at performing hill-climbing on $k$-NN graphs [17]. This work is the most related to ours; however, we introduce several significant differences. First, [17] starts to search from a point drawn randomly from a uniform distribution over the database. Following such a strategy is prone to reaching poor minima, especially for large and diverse databases. To circumvent this problem, [17] allows the system to restart the entire search with a new seed a pre-defined number of times, which is relatively computationally wasteful. We then propose to train a Random Forest adapted for retrieval that will very quickly generate seeds that are in general much closer to the minima reached after optimization (i.e. less edges are traversed). Thus we are able to reach bet-ter results, faster. Second, [17] only explores one path at a time, which is extremely greedy. Similar to bio-inspired techniques, we explore and 'optimize' multiple hypothesis at once. Again, this make the search less prone to getting stuck in bad local minima. All these elements contribute to a method that significantly outperforms [17].

## 3. Method Overview

The proposed method is essentially a discrete optimization technique that uses a hierarchical navigation graph and a retrieval forest. We build the graph using multiple distance measures in order to decrease the number of local minima, and use the reconstruction error (or a surrogate) when navigating through the graph. We discriminatively train trees to generate seeds to initiate the navigation in the first graph in the hierarchy. The search starts from multiple seeds, traverses the hierarchy of navigation graphs, and finally outputs a number of solutions which are used to initiate the continuous optimization step if refinement is needed.

A retrieval forest [13] outputs a list of samples for each tree, which are ranked based on how many times they were selected. The top-ranking vertices are used to initiate the search in the top level of the navigation graph.

At each iteration of the search, a candidate solution list is expanded by including all neighbors of the current estimates. The query image is compared against this list using the reconstruction error (or a surrogate measure), and the resulting energies used to rank the candidates. A few top-ranking vertices are then used to initiate the next iteration of the search in the same level of the graph. If the candidate solution set does not change throughout the iteration, or if a maximum number of iterations are met, we continue the search in the next level of the hierarchy, using the final list of vertices from the previous level as seeds. Once the search concludes in the bottom level, the solution is final and can optionally be sent to a continuous optimizer (e.g. PSO, gradient descent).

## 4. Retrieval Forests

A retrieval forest [13] is a randomized decision forest (RDF) [9], which acts as a hash function that assigns each query to a leaf. The leaves of a typical RDF store an empirical distribution over the samples. In retrieval forests, the leaves act as the entries of a lookup table and store the indices of dataset elements. During inference, each tree votes for a set of elements present in the training set, and we count the number of votes per element in order to rank them.

We use the standard greedy tree training algorithm to grow the trees. Each tree is trained with some randomness from the random generation of pairs of feature indices $\phi$ and thresholds $\tau$, and optionally also from bagging. We denote each pair $(\phi,\tau)$ as $\theta$, and the set of candidate random parameters $\theta$ in node $n$ as $\Theta_n$. Each node $n$ uses the randomly-generated $\Theta_n$ to greedily optimize

$$\theta_n^* = \underset{\theta \in \Theta_n}{\operatorname{argmax}} I_n(\theta), \qquad (1)$$

where $I_n$ is the information gain:

$$I_n(\theta) = E(\mathcal{S}_n) - \sum_{i \in \{\text{L,R}\}} \frac{|\mathcal{S}_n^i|}{|\mathcal{S}_n|} E(\mathcal{S}_n^i), \qquad (2)$$

Here, $E(\mathcal{S})$ is a measure of the differential entropy of the set $\mathcal{S}$ in descriptor space. Note that the left and right subsets $\mathcal{S}_n^i$ are implicitly conditioned on the candidate parameters $\theta$. As in [46], we use the determinant of the covariance matrix as the entropy measure.

Hash functions typically use dense projections, while we perform very sparse projections to traverse our trees more efficiently. The results obtained using the retrieval forest are then further refined in the next phases of the pipeline.

## 5. Multiscale Navigation Graph

The core of our discrete optimization method is the multiscale navigation graph, which allows for rapid refinement of a set of initial estimates in solution space. The initial seeds are provided by the retrieval forest (cf. Section 4), which are too crude to be used directly in a gradient-based continuous optimizer without a robust refinement procedure. The graph search rapidly refines these solutions while avoiding most local minima. Fig. 1 offers an intuition about the optimization procedure.

We start by constructing a multiscale graph $G = (V, E)$ from a set of samples, where the vertices $V$ are the individual samples and the directed edges $E$ are formed between vertices $p$ and $q$, if $q$ is among the $k$ nearest neighbors of $p$. The graph consists of multiple levels corresponding to different scales, each described with an adjacency list $G_i$, much like a pyramid. While top level holds a fraction of all samples, chosen uniformly, the bottom level keeps all the samples in the dataset. Every vertex in a higher level of the graph, must appear in every level below it. $k$ neighbors are calculated separately for each level.

Note that $E(p, q)$ does not necessarily imply $E(q, p)$ since nearest neighbor relationships are not symmetric, even if the distance metric is. The distance $D_m$ between samples $p$ and $q$ is denoted by $D_m(p, q)$, where $m$ denotes the metric used (e.g., Euclidean distance in a certain space).

The graph search algorithm is explained in Algo. 1. Given a query $t$, the search starts in the top level graph $G_T$ from an initial set of seeds $C$, which is provided by the retrieval forest as described in the previous section. A candidate list $C'$ is then formed from samples in $C$ and all their neighbors. For each candidate in $C'$, we measure their distance $D_t$ to the query. Note that the distance measure $D_m$ used to build $G$ is not necessarily the same as $D_t$. We rank these candidates based on their distances and choose $n$ samples to replace $C$. If $C$ does not change in a given iteration, or if a maximum number of iterations is reached, we switch to the next graph in the hierarchy, initiating a new search with the set $C$ where the previous graph converged to. The $C$ from the final level in the hierarchy is the output of the discrete optimization method.

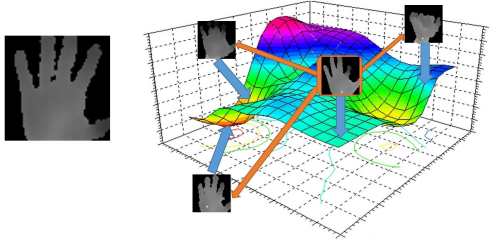

Figure 1. **Intuition for Navigation Graphs.** Left: query; right: reconstruction error as a function of the pose parameters. The current candidate pose is highlighted in orange; the optimization process consists of estimating whether any of its neighbor minimizes the reconstruction error any further. If this is the case, the optimization may consider them as potential descent directions. Orange arrows mark connections between candidates, and blue arrows denote their positions on the solution manifold.

## 6. Continuous Pose Refinement

The previous section has described how good camera hypotheses can be formed. Using these hypothesis, we follow [30, 11] and minimize the photo-consistency error in order to make accurate pose predictions. Given an initial pose hypothesis $p$ and a query image $C_q$, we optimize for the rigid body transform between $p$ and $C_q$ by minimizing the projection of the model of the scene $M$ to $C_q$. At any stage of the optimization, we raycast $M$ from the current pose of hypothesis $p$ and produce a color image $C_m$ and a depth image $D_m$. We then optimize the rigid transform $\mathcal{T}$ to maximize the photo-consistency when projecting $C_m$ into $C_q$ using $D_m$. Let $G$ denote the intensity image of $C_m$, $\pi_d$ the depth camera projection, and $\pi_c$ the color camera projection ($\pi_d$ and $\pi_c$ are defined by the respective camera intrinsics), then the photo-consistency error is given as

$$E_c(\mathcal{T}) = \sum_k^{\#\text{pixels}} \left\| G_q(\pi_c(\mathcal{T}\pi_d^{-1}(D_m(k)))) - G_m(k) \right\|_2^2.$$

Note that we discard pixels with invalid depth. The non-linear least-squares objective $E_c$ is solved using Gauss-Newton optimization. We use a hierarchical coarse-to-fine optimization strategy, with the finest hierarchy level containing the full color resolution, and each subsequent hierarchy level sub-sampled by a factor of 2.

Then, for query image $C_q$ and its top $k$ pose proposals $p_1, ..., p_k$, we minimize $E_c$ for each pose proposal, produc-

**Algorithm 1:** Graph traversal

**Inputs**: A new test sample $t$. $M$ adjacency lists denoted $G_m$, defining each level of the multiscale graph structure $\mathcal{G}$. A function $\mathcal{D}$ that estimates some distance between two images. A set $\mathcal{S}$ of seeds. A number of maximum iterations $it_{max}$. A desired number of predictions $K$. A maximum number of vertices to keep after each iteration $n$.

**Output**: $K$ approximate nearest neighbors of $t$

**Algorithm**:

$\mathcal{C} = \mathcal{S}$
**for** $lv = M$ *to* $1$ **do**
    **for** $it = 1$ *to* $it_{max}$ **do**
        $\mathcal{C}' = \mathcal{C}$
        **foreach** $c \in C$ **do**
            Using the neighborhood structure defined in $G_{lv}$, add all unvisited neighbors of $c$ in to $\mathcal{C}'$ without repetition, and mark them as visited
        **end**
        Sort $\mathcal{C}'$ according to $\mathcal{D}$ computed between $t$ and each element of $\mathcal{C}'$.
        $\mathcal{C} = $ first $n$ elements of $\mathcal{C}'$
        **if** $\mathcal{C}$ *has not changed this iteration* **then**
            break
        **end**
    **end**
**end**
**return** *first* $K$ *elements of* $\mathcal{C}$

ing potential refined poses $p'_1, ..., p'_k$. The final refined pose $p^*$ is then given by the pose $p'_i$ with $i = \arg\min_j E_c(p'_j)$. As this optimization is straightforward to parallelize, we can easily compute $\mathcal{T}$ on a GPU within only a few milliseconds for a given image pair.

## 7. Results

We evaluate our method on three important computer vision tasks. First, we show results on Hand Pose Estimation; second, we provide an evaluation on approximate nearest neighbor queries for Image Retrieval; third, we show results on RGB relocalization. Our method outperforms the baselines in all experiments, both in accuracy and speed. Open-source code for retrieval forests and Navigation Graphs will be released upon acceptance.

### 7.1. Experiments on Hand Pose Estimation

In this section, we first evaluate our method on the task of hand pose estimation using the synthetic dataset from [38]. This dataset comprises 100k training and 10k test hand poses that have been generated by rendering depth images of a synthetic hand model. For more details regarding the dataset generation, we refer the reader to [38].

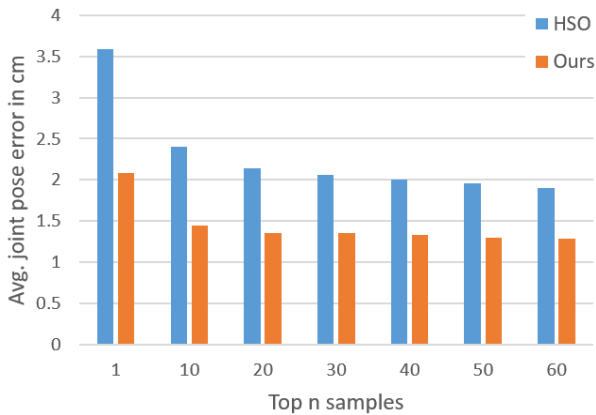

Figure 2. **Precision of hand pose predictions.** The abscissa defines how many seeds are used for any given hand image, and the ordinate represents the closest of those seeds in average joint error. These results are generated by considering the top $K$ samples from which we determine the lowest average joint pose error using an oracle. Note that our method not only provides more precise seeds than [41] but also extracts them an order of magnitude faster.

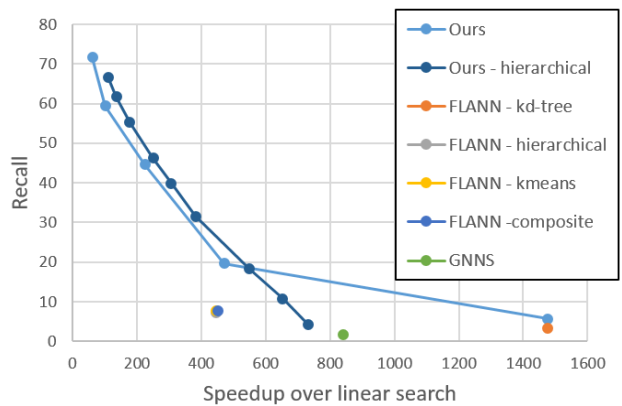

Figure 4. **Results on the GIST-1M dataset [21].** In these experiments, the task is to retrieve 100 neighbors, with recall computed against the actual 100 nearest neighbors. For our Multiscale Navigation Graph, we use a hierarchy of 2 graphs where the first contains 1/10th of the dataset and the second contains all of it. Our method outperforms all the baselines in FLANN in almost any regime. Note that the Retrieval Forest is not plotted on this figure, but reaches a speed-up over linear search of around 6500 for a recall of 2.07%.

In our experiments, we compare our results to the latest state of the art [41]. We follow [40] and use the average joint pose as an error metric. In Fig. 2, we show that regardless of the number of seeds we predict, our method is consistently better at predicting hand poses of lower average joint pose error than [41]. Furthermore, we extract our seeds an order of magnitude faster than [41].

We further compared our method on Dexter [40], NYU [44] and FingerPaint datasets [38] which only contain real images. In Fig. 3 we show that the proposed method significantly outperforms Retrieval Forest only and the reinitializer presented in [38]. We also demonstrate the power of our method under the presence of a powerful continuous LM based optimizer, where our method generates 10 seeds per frame to initiate the gradient descent. Each frame is treated independently (i.e. without tracking) to properly show the effect of using our reinitializer instead of the one presented in [38]. Our method shows significant gains on Dexter and NYU, and shows on par results for FingerPaint. The continuous optimization method and more experimental results can be found in [2].

### 7.2. Experiments on Image Retrieval

In this section, we demonstrate the performance of the proposed NN-search method on a well-established image retrieval dataset, GIST-1M [21].

In the experiments presented in Fig. 4, we compare our work against all the baseline methods implemented in the FLANN library [28] using the code provided on the authors' website as well as our own implementation of [17] (referred to as GNNS). Notably, our proposed method significantly outperforms all baselines from FLANN with a comparable compute budget. For instance, at $450\times$ speedup over linear search, the hierarchical, $k$-means, and composite baselines from FLANN only attain half the recall of our method. Additionally, our method is capable of reaching even higher recall rates if more compute budget is available. With a 5ms budget (on a laptop computer), we achieve recall rates over 70%, corresponding to a $60\times$ speedup compared to linear search. Note that to provide a fair comparison against [17], we set the number of restarts to be the same as the number of seeds we extract from the retrieval forest.

### 7.3. RGB Relocalization

**Dataset** The main dataset currently used for RGB and RGB-D relocalization is the 7-Scenes dataset [39, 15]. This dataset contains several scenes recorded from a KinectV1. These scenes are all very limited environments (at most $6m^3$), for which extremely precise results have already been reached [46]. A major increase in the volume in which to relocalize is a clear way to make the relocalization problem more challenging. Additionally, the 7-Scenes recordings suffer from low color quality (VGA resolution, motion blur artifacts, auto-white balance, auto-exposure) and no external calibration between the depth and RGB sensors. As these issues are all easily solvable hardware or calibration problems, they should not be part of benchmark evaluations

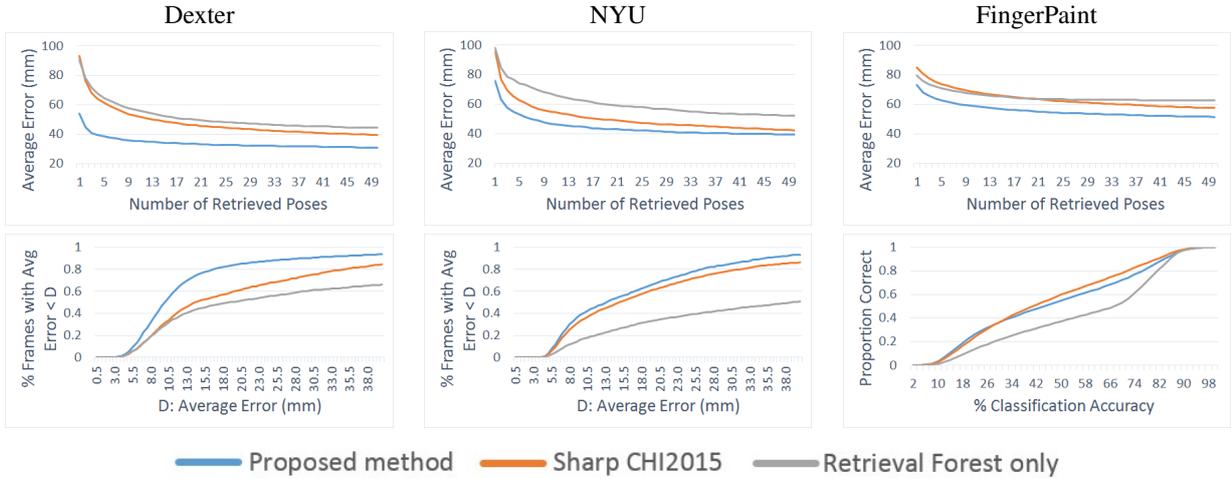

Figure 3. Hand pose estimation on real data. First row: pure pose regression. Second row: regressed poses optimized using the framework presented in [2]

of relocalization techniques. Fig. 6 illustrates these effects on the scene model quality, which is a major obstacle to synthesizing RGB frames consistent with the real world.

We thus introduce a new dataset which we believe will allow the community to push the boundaries of RGB and RGB-D relocalization further. Using a Structure.io depth sensor coupled with an iPad color camera, we capture various environments of significant physical extent (volumes an order of magnitude larger than those of 7-Scenes), as can be observed in Tab. 1. Note that both cameras have been calibrated and temporally synced. RGB image sequences are recorded at a resolution of $1296 \times 968$ pixels; the depth resolution is $640 \times 480$ and of similar quality to the KinectV1.

We capture four large RGB-D scenes, each of which is composed of several rooms as shown in Fig. 5. For each scene, two independent captures are performed. The first capture is used to reconstruct a scene model, with reconstruction performed using the VoxelHashing framework of [31] in combination with global bundle adjustment. This reconstruction provides the ground truth camera poses (6 d.o.f.) for each frame. We then uniformly sample a fixed number of poses around those ground truth camera poses and synthetically render the corresponding RGB images. This procedure is used to form the training set. The second recording is also reconstructed and bundle adjusted. Then, manual alignment between the two models is performed. The RGB images and camera poses of this second recording correspond to our test set.

Note that the dataset (RGB-D frames, ground-truth poses, and 3D models) will be made publicly available to the community.

**Parameter settings** The parameters of the system are held constant for all relocalization scenarios. The retrieval forest is comprised of 64 trees with a maximum depth of 13, where each decision node only uses 2 dimensions of the input vector to route the samples. For the Multiscale Navigation Graph, at any stage, the 20 closest neighbors of each candidate are considered and the 10 best candidates are passed to the next iteration over a maximum of 5 iterations. The 4 top candidates are then passed to a pose refinement step.

**Continuous Pose Refinement** Similarly to [46, 39] which perform RGB-D relocalization by first generating hy-pothesis and then optimizing them in a continuous fashion for high precision, we run a continuous pose optimizer on top of the top prediction from our method. To do so, we follow [30, 11] and minimize the photo-consistency error. It is worth noting that the energy surface is non-convex and hence good initial candidates are essential to sustain good relocalization rates. We refer the interested reader to the supplementary materials for more details. Note that precise camera relocalization is required in applications such as SLAM where it allows the system to recover from tracking failures.

| Sequence | Volume | # training frames | # test frames |
|---|---|---|---|
| Kitchen | $33m^3$ | 7160 | 357 |
| Living | $30m^3$ | 10554 | 493 |
| Bath | $24m^3$ | 22500 | 230 |
| Bed | $14m^3$ | 17800 | 244 |
| Kitchen | $21m^3$ | 14665 | 230 |
| Living | $42m^3$ | 34280 | 359 |
| Luke | $53m^3$ | 38382 | 624 |
| Floor 5a | $38m^3$ | 10680 | 497 |
| Floor 5b | $79m^3$ | 10360 | 415 |
| Copy | $26m^3$ | 13640 | 649 |
| Gates362 | $29m^3$ | 18641 | 386 |
| Gates381 | $44m^3$ | 20240 | 1053 |
| Lounge | $38m^3$ | 3160 | 327 |
| Manolis | $50m^3$ | 17940 | 807 |
| **TOTAL** | **$521m^3$** | **240002** | **6671** |

Table 1. **Physical description of our new dataset.** The recorded environments are significantly larger than the largest environment of the 7-Scenes dataset ($6m^3$).

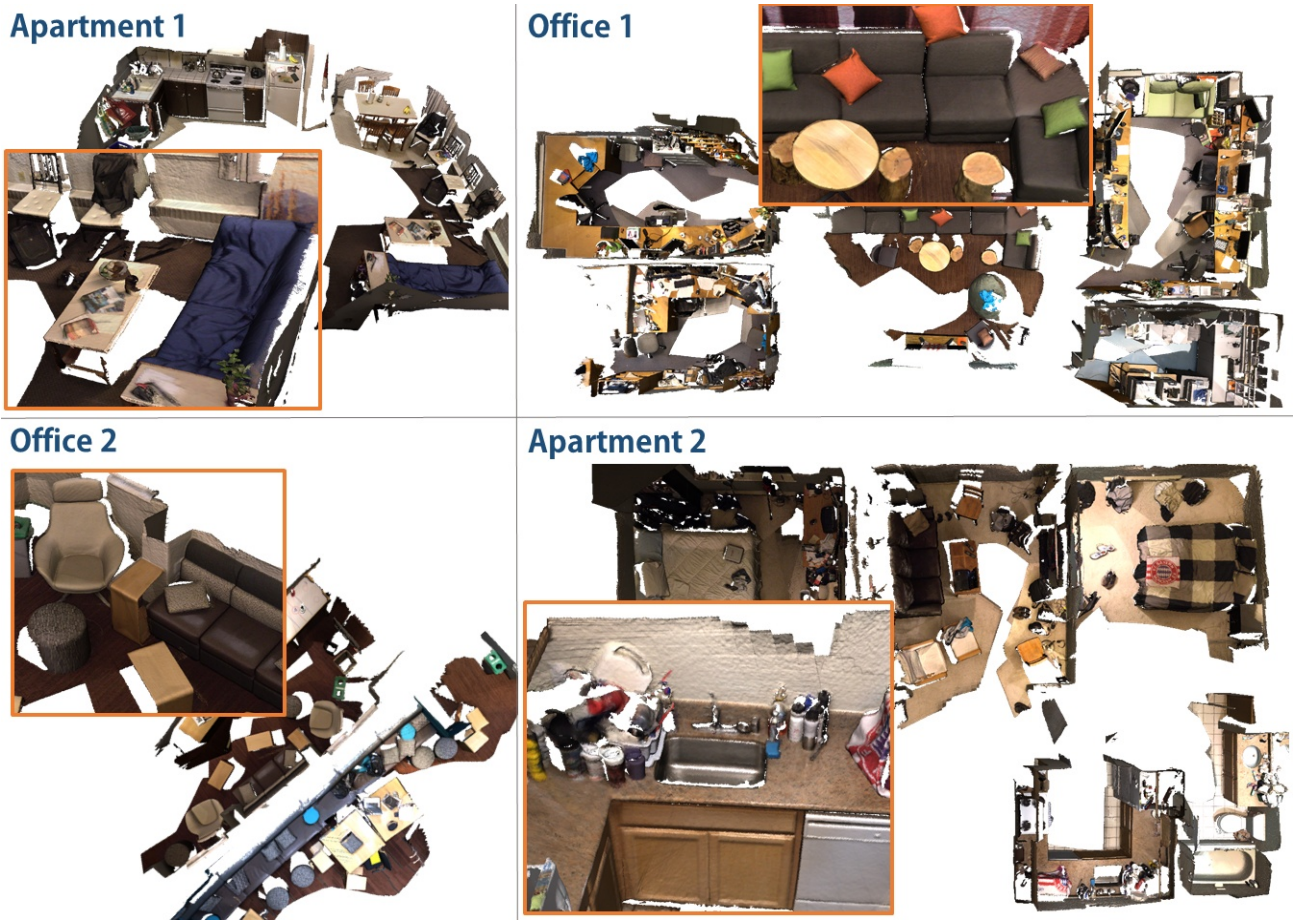

Figure 5. **A new public dataset for RGB and RGB-D camera relocalization.** Our new dataset comprises two apartments and two office scenes. There is a total of 14 rooms, each of which are individually larger than the largest scene of [39].

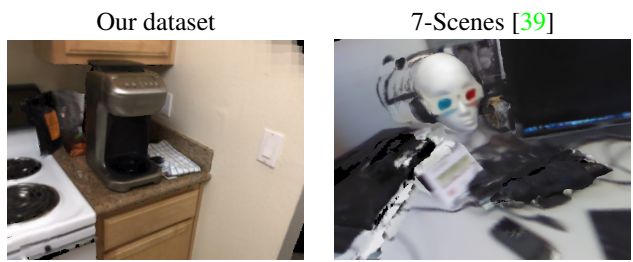

Figure 6. **Qualitative difference between our dataset models and those of [39].** Both images are raycast from their respective 3D models. Note that we are able to synthesize RGB frames of much higher quality.

**Speed** On a laptop equipped with an i7-4720HQ, passing a new sample down the entire test pipeline (retrieval forest and Multiscale Graph Navigation) takes less than 1ms. We run the pose refinement step on the GPU using an NVIDIA GTX Titan, which takes less than 10ms to fully optimize each pose proposal. For each query, we only optimize the top 4 pose proposals from the Multiscale Graph Navigation.

**Baseline comparisons** We compare our approach against two RGB relocalization baseline based upon matching sparse features, following mainstream feature-based relocalization approaches such as [47, 10, 29]. Those approaches either use the SIFT or the ORB algorithm for sparse feature extraction and description. First, sparse features are detected in the RGB-D training images, which are stored with their corresponding 3D positions and descriptors. At test time, descriptors are computed per query frame and matched. Triplets of strong feature matches are then used to generate candidate camera poses using the perspective 3-point method. These hypotheses are refined using a RANSAC optimization. Note that during training and test, at most 100 SIFT features or 1000 ORB features are extracted per image. In order to maximize the relocalization performacne of the baselines, we use brute force feature matching followed by RANSAC PnP from OpenCV's [6] implementation.

It is worth noting that we do not compare against PoseNet [22] as the only dataset for which they report results and we could potentially use is 7-Scenes [39]. As described earlier in the paper, the test relocalization pipeline for precise estimates requires a good calibration between the color and depth sensors, which is not the case in the 7-Scenes dataset where no external calibration has been provided. Nevertheless, it is worth noting that the results obtained by PoseNet are only slightly better than those of nearest neighbors. The best results from PoseNet requires 95ms on a GPU. Using this budget, we could fully optimize 115 cameras using the method presented in this paper. To get a feel for where the proposed method stands, we ran only the Retrieval Forest and Navigation Graphs on the 7-Scenes dataset. On average, we retrieve the nearest neighbor 97% of the time with an average run-time of 1ms (CPU).

**Main relocalization results**  Fig. 7 shows qualitative results obtained by the Multiscale Navigation Graph when fed with seeds generated by the retrieval forest. The first predictions are visually very similar to the queries. The predictions from Multiscale Navigation Graph often contain the same objects as the query, but sometimes from slightly different viewpoints. However, Fig. 8 shows that these predictions still lie in the convergence basin of the continuous pose refinement.

Table 2 lists the main quantitative results on our relocalization dataset. On average, the proposed system is able to relocalize 67.4% of the queries with error below 5cm and 5°. For applications that don't require this level of precision, our method scores 88% within 30cm and 10°.

Comparatively, the sparse relocalization baseline of SIFT matching and RANSAC PnP pose refinement results in an average relocalization rate of 51.8% within 5cm and 5°. Those results are obtained in ≈ 1500ms, whereas our Multiscale Navigation Graph followed by continuous pose optimization runs in ≈ 40ms. By replacing SIFT [27] with ORB [35], speed increases 3×, but accuracy falls 5-10%. Thus, we provide both better camera pose estimates and fast performance. We believe that these results would be very beneficial to many applications.

| Sequence | ORB + PnP | SIFT + PnP | Our method |
|---|---|---|---|
| Kitchen | 66.39% | 71.43% | **85.7%** |
| Living | 41.99% | 56.19% | **71.6%** |
| Bath | 53.91% | 48.70% | **92.2%** |
| Bed | 71.72% | **72.95%** | 66.4% |
| Kitchen | 63.91% | 71.74% | **76.7%** |
| Living | 45.40% | 56.19% | **66.6%** |
| Luke | 54.65% | 70.99% | **83.3%** |
| Floor 5a | 28.97% | 38.43% | **66.2%** |
| Floor 5b | 56.87% | 45.78% | **71.1%** |
| Copy | 43.45% | **62.40%** | 51.0% |
| Gates362 | 49.48% | **67.88%** | 51.8% |
| Gates381 | 43.87% | **62.77%** | 52.3% |
| Lounge | 61.16% | 58.72% | **64.2%** |
| Manolis | 60.10% | 72.86% | **76.0%** |
| **Average** | 52.99% | 61.56% | **67.4%** |

Table 2. **Main quantitative results for the task of RGB relocalization.** For each sequence, we show the percentage of refined poses within 5cm/5° for our method and generic baselines. Note that the baselines with ORB and SIFT are first optimized using a PnP optimization over RanSaC rounds. For SIFT + PnP, note that we use 100 features per image. For ORB + PnP, we use 1000 features per image. Also, note that the feature matching for SIFT and ORB is done exactly via brute-force matching.

However, despite the relatively good quality reconstructions which we obtain, all 3D models still remain inconsistent with the real world due to tracking errors in the reconstruction, as well as other artifacts such as motion blur. In fact, our method would be able to provide much more accurate results if the training and test images were consistent with each other; e.g., generated from the same model. We set up such an experiment, generating query images by raycasting the scanned scenes rather than using the original color frames captured by the iPad camera. In this case, we obtain significantly higher accuracy, as shown in Fig. 9. This suggests that there is room for improvement by using more accurate models; however, it is quite challenging to reconstruct large scenes at millimeter-level accuracy.

**Memory**  For all sequences, we require less than 25MB memory. Note that loading everything into RAM is not required and that lazy data access is possible if necessary.

## 8. Conclusion

In this work, we presented a novel discrete optimization method and demonstrated its applicability to several computer vision tasks, including RGB relocalization, hand pose estimation, and nearest neighbor queries for image retrieval. At the core of the pipeline lies our multiscale navigational graph. Given a list of seeds predicted by a random forest, the graph search quickly traverses the discrete manifold of solutions in order to make fast and accurate predictions. These predictions can then be further refined; for instance, in the case of RGB relocalization, we run a continuous pose optimization for precise 6DOF camera pose inference. The same applies to the results we have shown on hand pose estimation and image retrieval tasks.

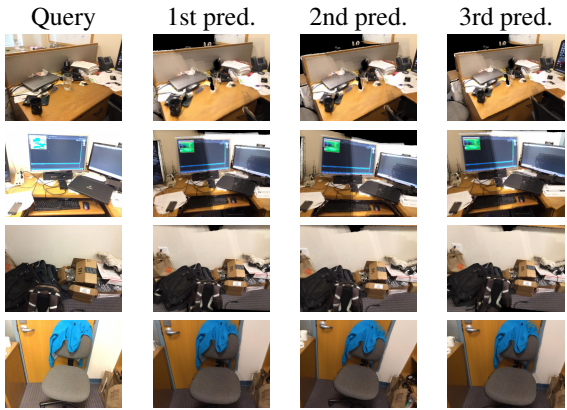

Figure 7. **Qualitative results from the Multiscale Navigation Graph.** Given a query image (left column) and seeds from the retrieval forest, the graph search produces viewpoints to the query. Our method is relatively robust even when some data is missing from the synthetic views or when the illumination changes.

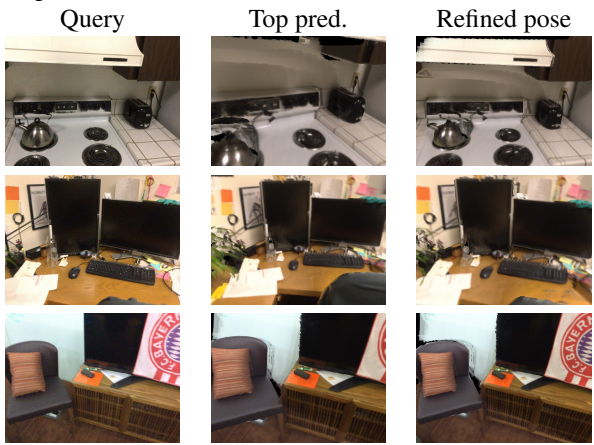

Figure 8. **Continuous pose refinement for predicted camera hypotheses.** From a query image (left column), the graph search searches for the most similar viewpoints. From the top 4 predicted viewpoints (the top prediction shown in the middle column), we optimize for the final pose by minimizing photometric error.

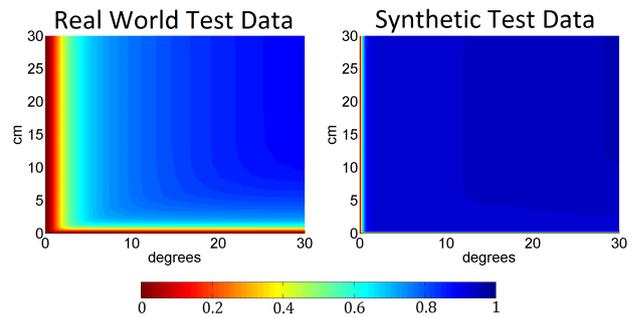

Figure 9. **Precision of the relocalization averaged across all scenes.** The value of the plot at $(x, y)$ represents the % of predicted cameras with translational and rotational error lower $x$ and $y$. Left: precision on real world test data. Right: precision on synthetic test data.